\documentclass[letterpaper, 10 pt, conference]{ieeeconf}  

\IEEEoverridecommandlockouts                              

\usepackage{epsfig} 
\usepackage{mathptmx} 
\usepackage{times} 
\usepackage{amsmath} 
\usepackage{amssymb}  

\usepackage{booktabs}
\usepackage{multirow}
\usepackage{url}

\usepackage{graphicx} 

\normalmarginpar 
\newcommand{\IEEEpreprintnotice}{
    \rotatebox{90}{\footnotesize This work has been submitted to the IEEE for possible publication. 
    Copyright may be transferred without notice, after which this version may no longer be accessible.}
}

\title{\LARGE \bf \vspace{-10pt}
Perceptual Motor Learning with Active Inference Framework for Robust Lateral Control\\
}

\author{Elahe~Delavari$^{1}$, John~Moore$^{2}$, Junho~Hong$^{3}$,~and~Jaerock~Kwon$^{4}$
\thanks{*This work was supported in part by the Ford/University of Michigan Research Alliance Program (10.13039/100007270) and in part by the National Science
Foundation (NSF) under Grant MRI 2214830.}
\thanks{$^{1}$E. Delavari, $^{3}$J. Hong, and $^{4}$J. Kwon are with the University of Michigan-Dearborn.
        {\tt\small \{elahed, jhwr, jrkwon\}@umich.edu} Corresponding author: J. Kwon.}%
\thanks{$^{2}$J. Moore is with Ford Motor Company.
        {\tt\small jmoor422@ford.com }}%
}

\begin{document}

\maketitle
\marginpar{\IEEEpreprintnotice} 

\thispagestyle{empty}
\pagestyle{empty}

\begin{abstract}
This paper presents a novel Perceptual Motor Learning (PML) framework integrated with Active Inference (AIF) to enhance lateral control in Highly Automated Vehicles (HAVs). PML, inspired by human motor learning, emphasizes the seamless integration of perception and action, enabling efficient decision-making in dynamic environments. Traditional autonomous driving approaches—including modular pipelines, imitation learning, and reinforcement learning—struggle with adaptability, generalization, and computational efficiency. In contrast, PML with AIF leverages a generative model to minimize prediction error (“surprise”) and actively shape vehicle control based on learned perceptual-motor representations. Our approach unifies deep learning with active inference principles, allowing HAVs to perform lane-keeping maneuvers with minimal data and without extensive retraining across different environments. Extensive experiments in the CARLA simulator demonstrate that PML with AIF enhances adaptability without increasing computational overhead while achieving performance comparable to conventional methods. These findings highlight the potential of PML-driven active inference as a robust alternative for real-world autonomous driving applications.

\end{abstract}

\section{INTRODUCTION}
The rapid evolution toward fully Highly Automated Vehicles (HAVs) demands decision-making frameworks that not only emulate human cognition but also robustly generalize across diverse and unforeseen environments without extensive retraining. Traditional approaches in autonomous driving—including Modular Pipelines (MP) \cite{Self-driving-cars:-A-survey}, Imitation Learning (IL) \cite{Alvinn}\cite{End-to-end-learning-for-self-driving-cars}, and Reinforcement Learning (RL) \cite{Learning-to-drive-in-a-day}—have paved the way for significant advances. However, each of these methods faces inherent limitations. MPs, while modular and interpretable, suffer from error propagation and inflexibility; IL is capable of mimicking expert behavior yet often fails when confronted with novel scenarios; and RL, despite its ability to learn through interaction, is heavily reliant on finely tuned reward-shaping, which is a process that is both labor-intensive and highly task-specific.

\begin{figure}[h!]
 \centering
    \includegraphics[width = 0.9\linewidth]{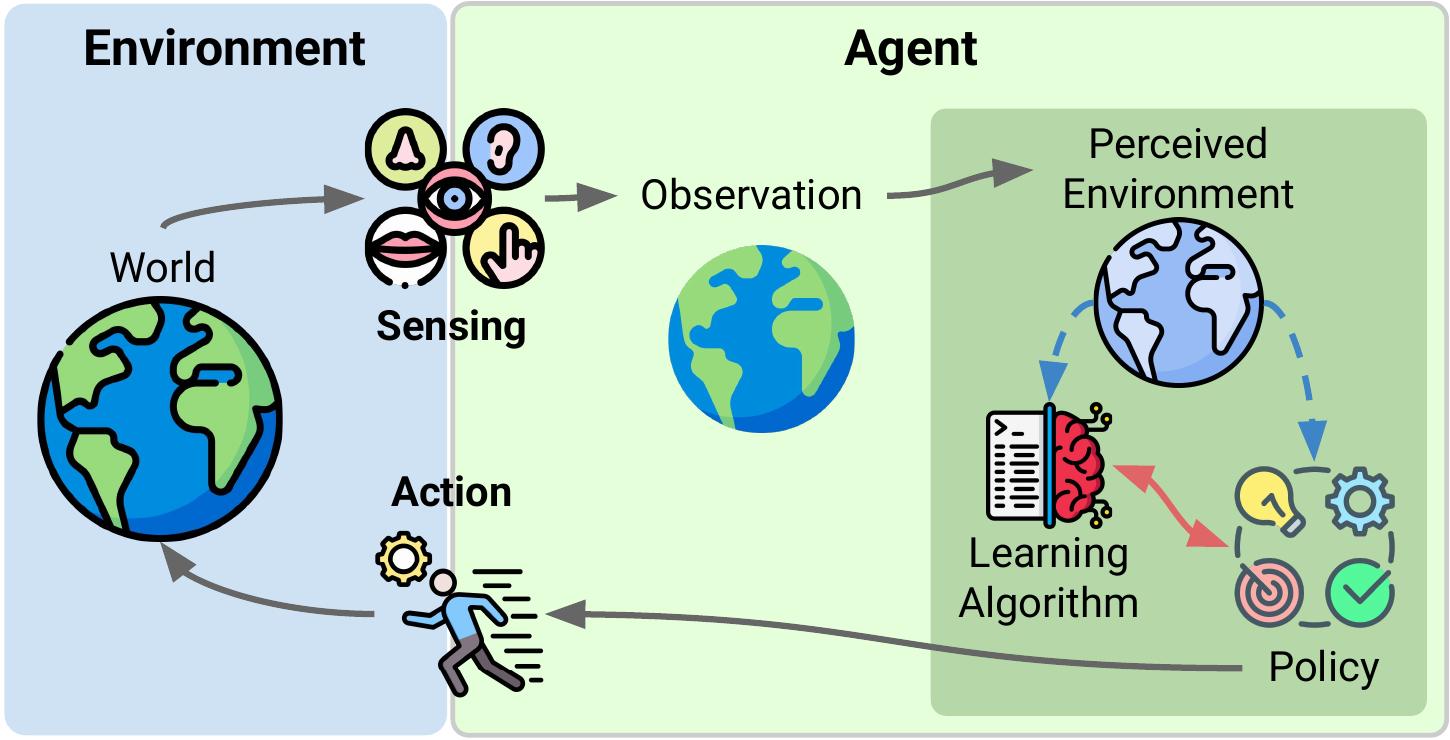}
 \caption{Overview of Perceptual Motor Learning (PML) with Active Inference Frameworks (AIF). The agent senses the environment through its sensors and makes observations. These observations shape the agent's beliefs and perception of the world. Based on its perception, the agent takes actions that, in turn, affect the real environment. (Icon made by Freepik from www.flaticon.com)}
 \label{fig:overview-pml}
\end{figure}

To address the aforementioned limitations, we propose Perceptual Motor Learning (PML)  \cite{Motor_Skill_Development}\cite{Perceptual_and_Motor_Development} with Active Inference Framework (AIF) \cite{Active-inference}. PML refers to the process by which sensory inputs (perception) and motor outputs (action) become integrated to enable effective interaction with the environment. In early human development, this integration is crucial: while fundamental movement skills (e.g., hopping, jumping, running, and balance) lay the groundwork, perceptual motor development connects these skills with sensory processing, enabling children to develop body awareness, spatial awareness, directional awareness, and temporal awareness \cite{A_Developmental_Perspective}. 
To realize a PML, we propose to use AIF, which has emerged as a promising alternative to conventional control theories, grounded in theories of human brain function that posit the continuous prediction of sensory inputs and the minimization of discrepancies between expected and actual observations \cite{Free-energy-and-the-brain}. By unifying perception and action within a single generative framework, active inference allows an autonomous agent to actively anticipate environmental changes rather than merely reacting to them. Nevertheless, our work extends this paradigm by incorporating principles from perceptual motor development—a concept traditionally rooted in developmental psychology. Fig.~\ref{fig:overview-pml} shows an overview of our method.

In our framework for autonomous driving, we draw a parallel to this human developmental process. Our model leverages a deep learning architecture to fuse perceptual and motor information, constructing an internal world model that captures the causal dynamics of the environment. This model enables the system to predict the consequences of its actions and adapt to new road conditions without retraining—addressing the core issue of generalization that plagues traditional IL and RL methods. Rather than blindly imitating expert behavior or relying on externally defined rewards, our approach infers \textit{how the world works} to support robust and flexible decision-making.

To evaluate the effectiveness of our approach, we applied it to the lateral control of self-driving vehicles in a simulated urban environment. The model was trained on data from a simulated large urban area with multiple-lane highways and tested in distinctively different settings: a town with mountainous roads and a small town characterized by multiple bridges. This diverse testing environment highlights the model’s ability to generalize across varied road structures. 

In summary, the key contributions of our work are:

\begin{itemize}
    \item To our best knowledge, this study is the first integration of PML with AIF in HAVs. By leveraging an internal world model to learn the causal relationships between actions and environmental changes, our system adapts to new scenarios without requiring retraining, ensuring robust performance across diverse environments.  
    \item PML with AIF allows greater resilience and adaptability in novel driving conditions, whereas traditional IL methods suffer from long tail problems.  
    \item PML with AIF streamlines training, making it both computationally efficient and easily scalable across different driving tasks by eliminating the need for complex reward engineering.  
\end{itemize}

\section{Related Work}

There are different methods for controlling an HAV in an environment. One of the most popular methods is MP in autonomous driving systems, which generally consists of dedicated modules for perception, prediction, planning, and control. While this approach offers clear advantages in terms of interpretability and modularity, it also has several limitations. 
Errors in upstream modules (e.g., perception) can propagate downstream, leading to compounding errors in prediction, planning, and control. 
MPs can be inflexible, making it difficult to adapt the system to new tasks or environments without significant re-engineering. This rigidity can limit the scalability and versatility of the system. 
Joint optimization across modules is challenging due to the different nature of tasks each module performs \cite{End-to-end-driving-via-conditional-imitation-learning}\cite{Learning-by-cheating}.

Another method that is vastly used is Imitation Learning (IL) which has been a popular method for developing autonomous driving systems due to its ability to mimic expert behavior from demonstrations and a lot of work has been done in this domain \cite{Alvinn}\cite{End-to-end-learning-for-self-driving-cars}\cite{End-to-end-driving-via-conditional-imitation-learning}\cite{End-to-end-learning-of-driving-models-from-large-scale-video-datasets}. Traditional IL methods, however, often struggle with generalizing to unseen states not covered by the expert's demonstrations. When the agent encounters scenarios outside the training data, it may fail to perform adequately, leading to safety and reliability issues \cite{End-to-end-driving-via-conditional-imitation-learning}\cite{Active-inference-integrated-with-imitation-learning}. 
Additionally, in complex scenarios such as urban driving, high-level decisions (e.g. choosing to turn left or right at an intersection) cannot be easily inferred from perceptual input alone, leading to ambiguity and oscillations in the agent’s behavior. To overcome this, Codevilla et al.\cite{End-to-end-driving-via-conditional-imitation-learning} introduced Conditional Imitation Learning (CIL), where the driving policy is conditioned on high-level commands. 

Direct perception methods are also the other category of methods that are used \cite{Chen_2015_ICCV}. Sauer et al. \cite{Conditional-affordance-learning-for-driving-in-urban-environments} propose Conditional Affordance Learning (CAL), a direct perception approach for autonomous urban driving. This method maps video inputs to intermediate representations (affordances) that are then used by a control algorithm to maneuver the vehicle. CAL aims to combine the advantages of MP and IL by using a Deep Neural Network (DNN) to learn low-dimensional, yet informative, affordances from high-level directional inputs. The approach handles complex urban driving scenarios, including navigating intersections, obeying traffic lights, and following speed limits. However, choosing the best set of affordances is challenging. Furthermore, the effectiveness of this method relies heavily on precise affordance predictions. Any inaccuracies or errors in predicting affordances may result in less than optimal or even hazardous driving decisions.

Use of RL has shown promising results in training autonomous agents through trial and error\cite{Learning-to-drive-in-a-day}\cite{End-to-end-model-free-reinforcement-learning-for-urban-driving-using-implicit-affordances}\cite{Cirl}. However, it also has its own set of limitations. RL methods typically necessitate extensive sample data to develop effective policies, which can be especially challenging in real-world settings where data acquisition is costly and time-intensive. Achieving a balance between exploration (testing new actions) and exploitation (leveraging known actions) presents a substantial hurdle in RL. 
Designing appropriate reward functions is critical for RL but can be difficult and problem-specific. Poorly designed rewards can lead to unintended behaviors or sub-optimal performance. 
Ensuring the stability and convergence of RL algorithms is challenging, especially in complex and dynamic environments. Instabilities during training can lead to unreliable and unpredictable agent behavior \cite{End-to-end-driving-via-conditional-imitation-learning}\cite{Learning-by-cheating}.

Liang et al.\cite{Cirl} introduce Controllable Imitative Reinforcement Learning (CIRL), which combines IL and RL to address the challenges of autonomous urban driving in complex, dynamic environments. CIRL utilizes a Deep Deterministic Policy Gradient (DDPG) approach to guide exploration in a constrained action space informed by human demonstrations. This method allows the driving agent to learn adaptive policies tailored to different control commands (e.g., follow, turn left, turn right, go straight). Despite using IL to guide initial exploration, CIRL still faces challenges associated with exploration in RL. The agent may struggle to discover optimal policies in highly dynamic and unpredictable environments without effective exploration strategies.

Active inference has been gaining interest in HAV research. Friston et al. \cite{The-anatomy-of-choice} proposed it as a process theory of the brain that minimizes prediction error (\textit{surprise}) by updating its generative model or inferring actions that reduce uncertainty. While applied to simple simulated agents, its use in realistic scenarios has been limited. Catal et al. \cite{Deep-active-inference-for-autonomous-robot-navigation} integrated deep learning with active inference for real-world robot navigation, using high-dimensional sensory data to construct generative models without predefined state spaces. Their work demonstrated successful navigation in a warehouse setting, though in a controlled environment.
de Tinguy et al. \cite{Spatial-and-Temporal-Hierarchy-for-Autonomous-Navigation} introduced a scalable hierarchical model for autonomous navigation, combining curiosity-driven exploration with goal-directed planning using visual and motion perception. Their multi-layered approach (context, place, motion) facilitated efficient navigation with fewer steps than other models, though validation was limited to simulations in a mini-grid environment.
Nozari et al. \cite{Active-inference-integrated-with-imitation-learning} combined active inference with IL to overcome IL's limitations. Their framework employs a Dynamic Bayesian Network (DBN) to encode expert demonstrations and includes an active learning phase for policy refinement. This approach dynamically balances exploration and exploitation, enhancing adaptability. However, DBNs impose high computational overhead, making them impractical for resource-constrained autonomous driving platforms.


To the best of our knowledge, most prior studies on action-based future scene prediction focus on standard driving scenarios \cite{Predicting-future-frames-using-retrospective-cycle-gan}\cite{Practical-Issues-of-Action-Conditioned-Next-Image-Prediction} or simplified environments like Atari games \cite{Action-conditional-video-prediction-using-deep-networks-in-atari-games}\cite{Deep-action-conditional-neural-network-for-frame-prediction-in-Atari-games}, where predicted images closely resemble the input. Even in these cases, models often struggle to generate high-quality predictions, leading to blurriness. Unlike previous studies that focus on predicting future scenes in simple or highly constrained environments, we apply this idea within a PML with AIF, integrating an encoder-decoder DNN architecture \cite{Xception} as the forward transition model. Our approach explicitly models the effect of steering actions on future observations. 
By leveraging a generative model within AIF, we improve scene prediction, leading to more reliable action selection and decision-making under uncertainty.

\section{Method}

Our approach integrates PML with AIF to develop a computationally efficient and explainable autonomous driving system. PML allows the agent to learn from sensory experiences, dynamically linking perception with motor execution, while active inference ensures that decisions minimize prediction errors based on the Free Energy Principle (FEP)\cite{The-free-energy-principle}. In this work, decision-making is guided by the vehicle's current state, sensory observations from a single front-facing camera, an internal world model, and predictions of how the environment will change based on potential actions. Notably, our model operates solely on visual perception, without relying on privileged information such as LiDAR, GPS, or HD maps. This design choice ensures that the model learns and generalizes from vision alone, making it more adaptable and comparable to human driving capabilities.

\subsection{Perceptual Motor Learning with Active Inference}
In autonomous systems, PML enables agents to build internal representations of the environment and adapt their actions dynamically based on sensory feedback. By leveraging PML principles, an autonomous vehicle can enhance its control strategies through learned associations between environmental perception and motor execution. This forms the foundation for our approach, where we integrate PML within an AIF to optimize decision-making under uncertainty.
According to FEP, agents minimize Free Energy (FE) \cite{The-free-energy-principle}, a measure of surprise or prediction error. However, since FE is intractable to compute directly, an approximation called Variational Free Energy (VFE) \cite{Active-inference} is used. VFE enables Bayesian inference, allowing an agent to infer hidden states from sensory observations. By minimizing VFE, the agent continuously updates its beliefs, ensuring its internal model remains consistent with sensory input.
In our case, input images serve as observations, while the states remain hidden. Consequently, we model the environment as a Partially Observable Markov Decision Process (POMDP), where the agent must infer hidden states from sensory data.
The key variables and functions in our framework are defined as follows:

\begin{itemize}
    \item \textbf{Variables:}
    \begin{itemize}
        \item Hidden state: $s_t \in \mathbb{R}^{d_S}$
        \item Action: $a_t \in \mathbb{R}^{d_A}$
        \item Observation: $o_t \in \mathbb{R}^{d_O}$
        \item Data: $D = \{o_t, a_t, o_{t+1}\}_{t=0}^T$
    \end{itemize}

    \item \textbf{Functions:}
    \begin{itemize}
        \item Observation model: $\hat{o}_t = f_o(s_t)$
        \item Forward transition model: $\hat{o}_{t+1} = f_s(o_t, a_t)$
        \item Inverse transition model: $\hat{a}_t = f_s^{-1}(o_t, o_{t+1})$
        \item Distance model: $d_t = f_d(o_t, \hat{o}_t)$
        \item Policy: $\pi: s_t \to a_t$
    \end{itemize}
\end{itemize}

The VFE serves as an upper bound on surprise and is used to infer hidden states:

\begin{equation}
\begin{aligned}
    F = \mathbb{E}_{q(s_{t+1})} \Big[ & -\ln p(o_{t+1} | s_{t+1}) \Big] \\
    & + D_{KL} \big(q(s_{t+1}) \parallel p(s_{t+1} | o_t)\big),
\end{aligned}
\end{equation}

where \( p(o_{t+1} | s_{t+1}) \) is the likelihood of the next observation given the inferred hidden state, \( p(s_{t+1} | o_t) \) is the prior belief about the next state given past observations, and \( D_{KL} \big(q(s_{t+1}) \parallel p(s_{t+1} | o_t)\big) \) measures the divergence between the approximate posterior and the prior belief, ensuring the update remains close to the true distribution.


To implement PML with AIF for autonomous driving, we employ a two-stage process consisting of offline \textit{task-agnostic} training and online action selection. In the offline phase, we pretrain a forward transition model  \( f_s \) \cite{An-internal-model-for-sensorimotor-integration} to learn transition dynamics, enabling the agent to predict how future observations evolve under different actions. Once trained, this model remains fixed and does not require further updates. In the online \textit{task-specific} phase, the agent predicts future states for all possible actions at each timestep and selects the action that minimizes the EFE to optimize decision-making under uncertainty.

In the context of autonomous driving, AIF involves the vehicle continuously predicting its sensory inputs, which consist of visual observations of the road, and minimizing prediction error through either perception, by updating its internal model, or action, by making steering adjustments. To model perception, We employ a forward transition model to predict future observations given the current state and action. Meanwhile, action selection is performed through sensory-motor simulation, which effectively serves as an inverse transition model \( f_s^{-1} \) by inferring the action that would lead to a desired future state. In this process, the agent evaluates covert actions before execution, selecting the one that minimizes the EFE. Given that our task focuses on lane-keeping, we incorporate \textit{preference}-based  \cite{Contrastive-active-inference} action selection, encouraging actions that maintain lane alignment.

\subsection{Perception Model}
\begin{figure*}
\vspace{1mm}
\centering
   \includegraphics[width = 0.85\textwidth]{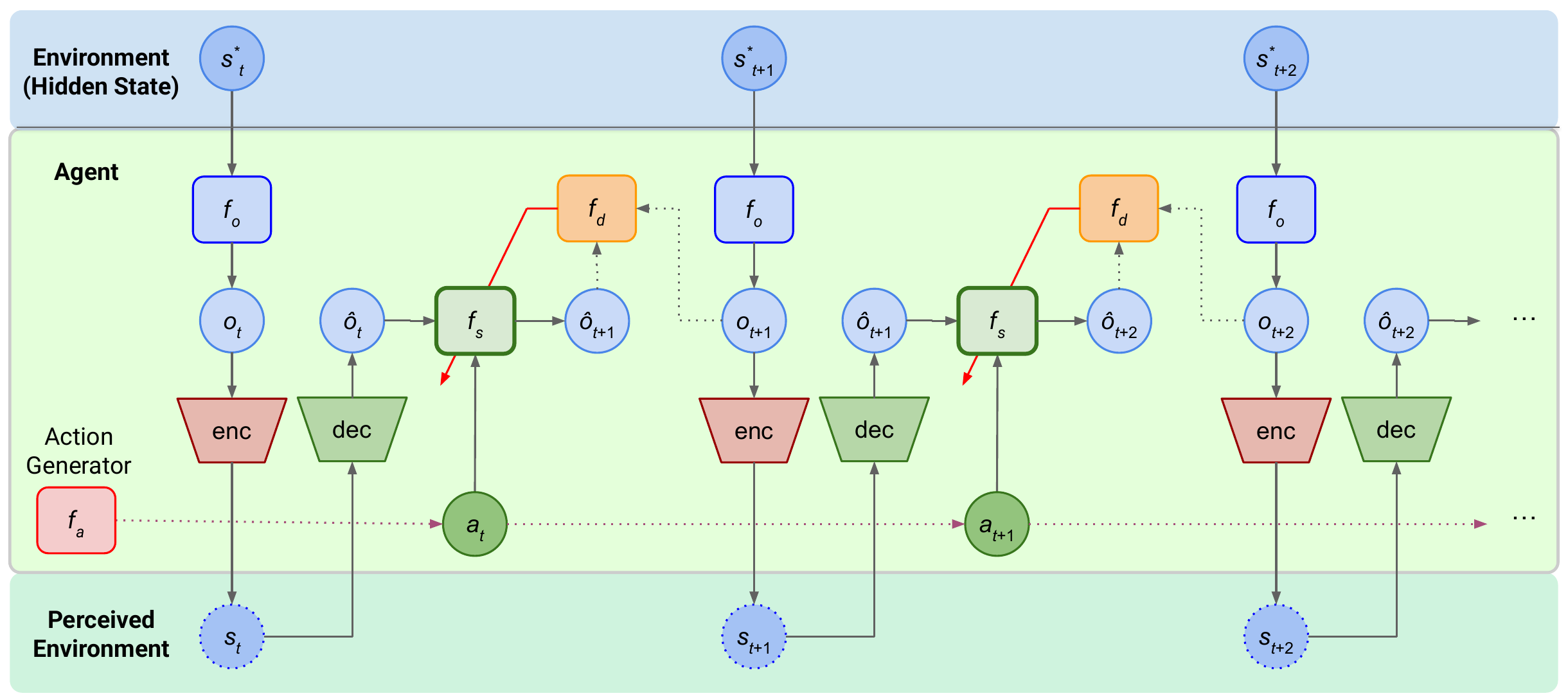}
\caption{
Task-agnostic offline perception model training. A hidden state \(s^*_t\) is perceived through the observation model \(f_o\). An observation \(o_t\) is encoded and kept inside the agent as a perceived state. When an action \(a_t\) is applied to the environment by the agent, the causal changes in the environment should be observed as \(\hat{o}_{t+1}\). The forward transition model \(f_s\) is being trained using the distance model \(f_d\) with \(o_{t+1}\) and \(\hat{o}_{t+1}\). The encoder and decoder can be the inside of \(f_s\) in practice. }
\label{fig:perception-model-training}
\end{figure*}

For perception, a forward transition model \( f_s \) is trained to contain the necessary information for the agent to understand the environment in the context of the driving task. This \( f_s \) is used to generate future images influenced by the steering angle. Training process of the forward transition model \( f_s \) is depicted in Fig.~\ref{fig:perception-model-training}.
For learning a forward transition model \( f_s \), this paper uses a method inspired by action-based representation learning \cite{Action-based-representation-learning-for-autonomous-driving}.

As human beings, it is simple to predict how the environment will change based on the change of steering action, as humans are exceptionally good at generating missing information. 
However, it is not straightforward for a feed-forward DNN to predict the details of a scene when provided only one frame and the corresponding action—especially when the future scene is significantly different from the current scene. The primary challenge lies in the network’s difficulty in learning a strong correlation between the current and future scene, even with the help of action.

We used a U-Net with an Xception-style architecture \cite{Xception}, incorporating task-specific modifications to better capture the action and observation dependency. U-Net \cite{U-net} is well-suited for tasks with limited training data, as it efficiently extracts hierarchical features using convolutional layers in both the encoding and decoding paths. Its skip connections help preserve spatial information from early layers, enhancing feature retention. These properties make it an ideal choice for the forward transition model.

\subsection{Action Selection Model}
Minimizing VFE ensures that the agent's internal beliefs align with observations, allowing it to infer states accurately. While VFE is crucial for state estimation, it does not dictate which actions the agent should take. For decision-making, the agent minimizes Expected Free Energy (EFE) \cite{Active-inference}, which predicts the impact of different actions on future states.
To make goal-directed decisions, the agent minimizes the EFE, which considers both epistemic (information-seeking) and pragmatic (goal-aligned) components:

\begin{equation}
\begin{aligned}
    G = \mathbb{E}_{q(o_{t+1}, s_{t+1} | a_t)} \Big[ & D_{KL} (q(s_{t+1} | a_t) \parallel p_{\text{pref}}(s_{t+1})) \\
    & - H(q(o_{t+1} | s_{t+1}, a_t)) \Big],
\end{aligned}
\label{eq:efe}
\end{equation}
where \( D_{KL} (q(s_{t+1} | a_t) \parallel p_{\text{pref}}(s_{t+1})) \) encourages the selection of actions that lead to desired states, and \( H(q(o_{t+1} | s_{t+1}, a_t)) \) represents uncertainty reduction, guiding exploration.

\begin{figure}
\centering
   \includegraphics[width = 0.85\linewidth]{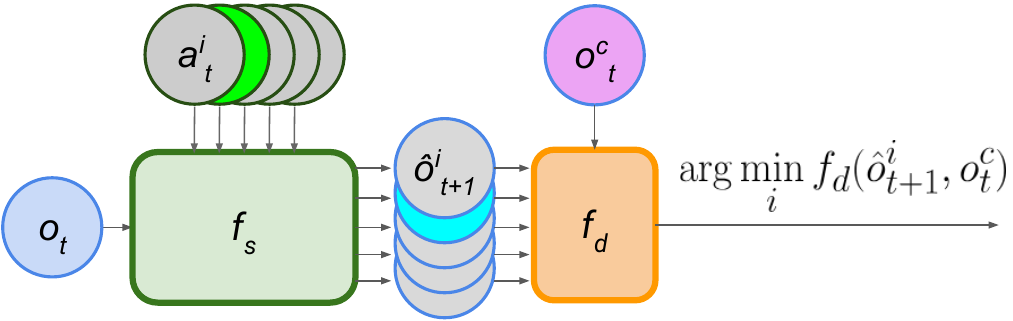}
\caption{Action selection. The forward transition model \(f_s\) generates causal observations based on covert actions. \(o^c_t\) is \textit{preference} for a task. The distance model \(f_d\) calculates the dissimilarity between the observations and the \textit{preference} and finds which action will generate the most similar observations to the \textit{preference}. }
\label{fig:action-selection}
\end{figure}

The agent balances goal-directed behavior with learning from its environment by seeking smaller \( G\). In our work, we focused on minimizing the first part of EFE with action selection, as the exploration is not beneficial for our case due to offline training of the forward transition model \( f_s \). 
This approach allows the vehicle to adapt its driving behavior in real-time, making steering adjustments that minimize prediction errors. 
To find an action that minimizes \(G\), the agent generates covert actions to imagine future states caused by the actions.
To align with goal-directed behavior, we compare predicted observations \( \hat{o}_{t+1} \) with a preferred future sensory state \( o_{\text{pref}} \) using Structural Similarity Index Measure (SSIM) \cite{Image-quality-assessment}. The SSIM index is between -1 and 1, where 1 is perfect similarity, and  -1 is maximum dissimilarity. Thus, our distance model (\(f_d\)) can be defined as \(1-\text{SSIM}\). The action can be selected by \eqref{eq:action-selection}. Fig~\ref{fig:action-selection} illustrates the action selection process.

\begin{equation}
\arg\min_{i} (1-\text{SSIM} (\hat{o}_{t+1}, o_{\text{pref}})).
\label{eq:action-selection}
\end{equation}

\begin{figure}
\centering
   \includegraphics[width = 0.49\linewidth]{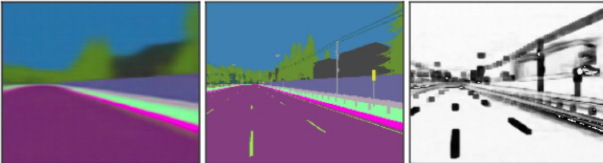}
   \includegraphics[width = 0.49\linewidth]{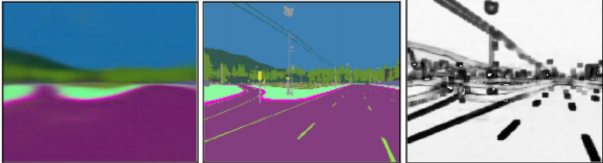}
\caption{Predicted and the ground truth image for semantic segmented images with the SSIM difference between them.}
\label{fig:ssim-examples}
\end{figure}

The preferred future sensory state is represented as \textit{preference}. For instance, in a lane-keeping scenario, it corresponds to a straight road within the same lane.
Fig.~\ref{fig:ssim-examples} illustrates the predicted semantic segmented images alongside their ground truth, highlighting the SSIM difference. Road only grayscaled version of the semantic segmented image used to simplify the problem and reduce computational overhead while maintaining the core driving information.

In this paper, we define steering as the action, while speed is kept constant to eliminate extra complexity. The future scene is predicted based on the current scene and the selected action. The forward transition model \( f_s \) provides prior knowledge of how different steering actions influence the future scene for the HAV. 
To guide decision-making, the model also requires an expectation of the future—referred to as \textit{preference}. 
Using the current state and the transition model \( f_s \), the HAV selects an action by comparing predicted future scenes with the expected \textit{preference}. The action associated with the highest similarity to this expected outcome is chosen.

\section{Experimental Setup}

\subsection{DNN Architecture and Training}
\subsubsection{PML}
We used a U-Net Xception-style model in \texttt{Keras 3} \cite{keras-link}. The input image was encoded with the encoding part of the U-Net and then the action was added by concatenating the action through Dense layers. For this task, the upsampling layers are changed to the conv2d\_transpose layer as they can make the reconstructed image less blurry. In addition, as only the road part of the image is considered, grayscale images are used. So, one channel in both the input layer and output layer is used. The input and output images have the same size of  \( 160 \times 160 \times 1 \). The RELU activation function is used for encoding and decoding layers and the ELU activation function is used for the part where the action was added. For the output layer, the sigmoid activation function was used. A learning rate of 1e-4 and batch size of 128 were used to train this model.

We tested the trained model in the CARLA (Car Learning to Act) \cite{CARLA} simulator using a covert action approach to find an optimal steering command. The model predicted the future image for twenty different steering values between -1 and 1, with a step size of 0.1, and the predicted images were compared with a \textit{preference} for the task. The comparison was conducted using SSIM, selecting the action corresponding to the predicted image with the highest SSIM score. This process ensured that for each input image, the optimal steering action was chosen, enabling the car to drive in the simulator. The fourth future image prediction at a constant speed was used for driving in the experiments.

\subsubsection{Imitation Learning}
Imitation learning encompasses various approaches, including Behavioral Cloning (BC) and Inverse Reinforcement Learning (IRL). To validate the PML agent's adaptability to new scenarios, we used BC for comparison, training a model to replicate expert demonstrations through supervised learning and evaluating their online driving performance.
The DNN architecture of the BC agent is shown in Table \ref{table:nn_architecture}. A learning rate of \(10^{-3}\) and batch size of 64 were used to train this DNN. 

\begin{table}[h!]
\centering
\caption{DNN architecture used for the BC agent}
\label{table:nn_architecture}
\begin{tabular}{@{}p{1.8cm}p{2.5cm}p{1.5cm}p{1.5cm}@{}}
\toprule
\textbf{Layer Type} & \textbf{Description} & \textbf{Kernel Size / Units} & \textbf{Activation} \\ \midrule
Input & Input Image & (160, 160, 3) & - \\ \midrule
Conv2D + BN & 32 filters, stride 2 & 5x5 & ReLU \\ \midrule
Conv2D + BN & 64 filters, stride 2 & 3x3 & ReLU \\ \midrule
Conv2D + BN & 128 filters, stride 2 & 3x3 & ReLU \\ 
Dropout & 20\% rate & - & - \\ \midrule
Conv2D + BN & 256 filters, stride 2 & 3x3 & ReLU \\ \midrule
Flatten & Flatten & - & - \\ \midrule
Dense & Fully connected & 512 & ReLU \\ 
Dropout & 20\% rate & - & - \\ 
Dense & Fully connected & 128 & ReLU \\ 
Dropout & 20\% rate & - & - \\ 
Dense & Fully connected & 64 & ReLU \\ 
Dense & Fully connected & 1 & - \\ \bottomrule
\end{tabular}
\end{table}

\subsection{Simulator and Dataset}
CARLA is a high-fidelity simulator that provides a realistic environment for developing, training, and validating autonomous driving technologies in urban settings. CARLA has multiple versions, each supporting different towns and features. 
We used CARLA 0.9.14 to test and validate the proposed PML with AIF in terms of online driving performance. Then, we also used CARLA 0.8.4 to compare other previously published methods in the original CARLA benchmark (referred to as CoRL2017) because the CoRL2017 benchmark is only compatible with the 0.8.4 version.

\begin{figure*}
\vspace{5mm}
\centering
   \includegraphics[width=.30\textwidth]{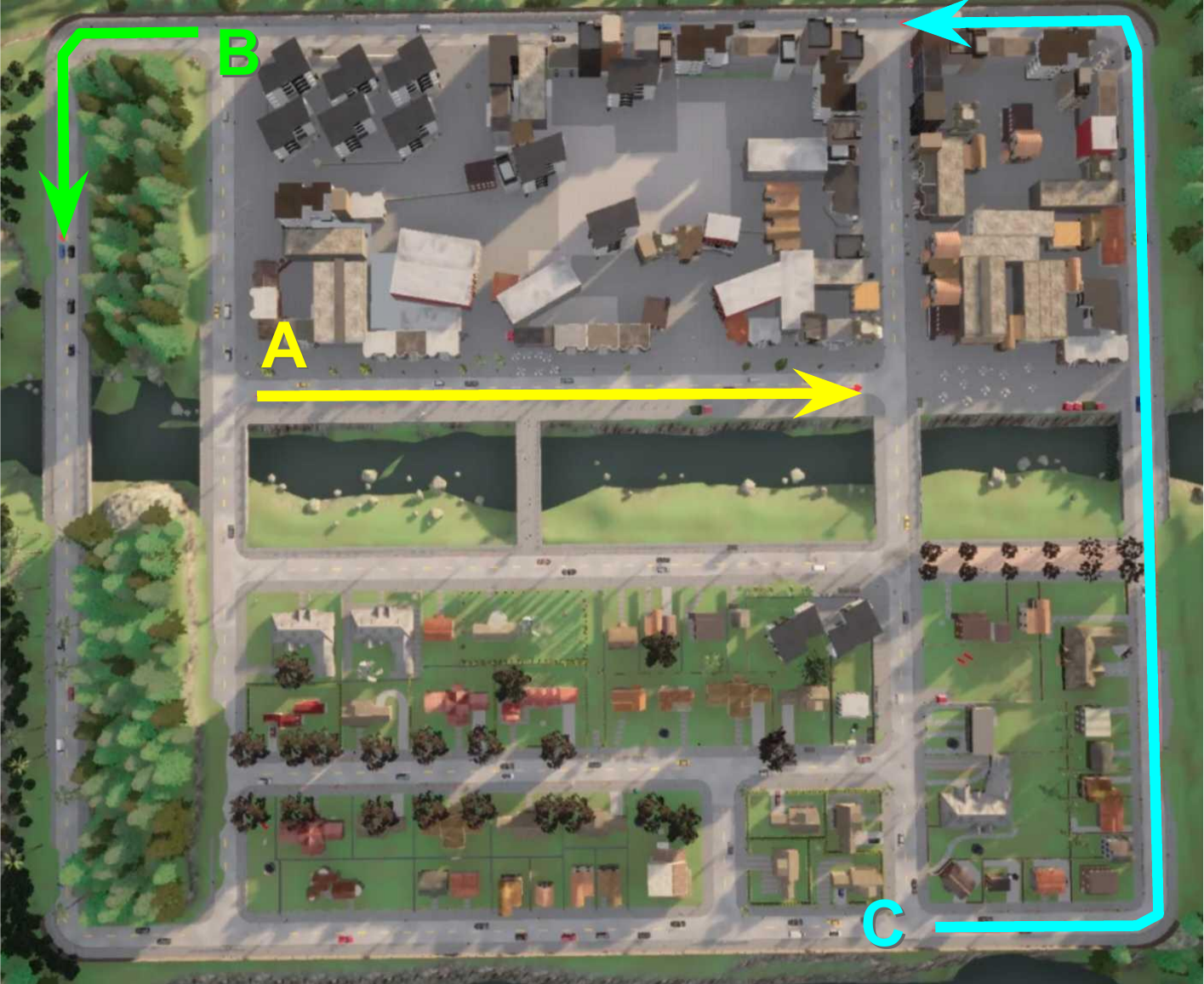}
   \includegraphics[width=.225\textwidth]{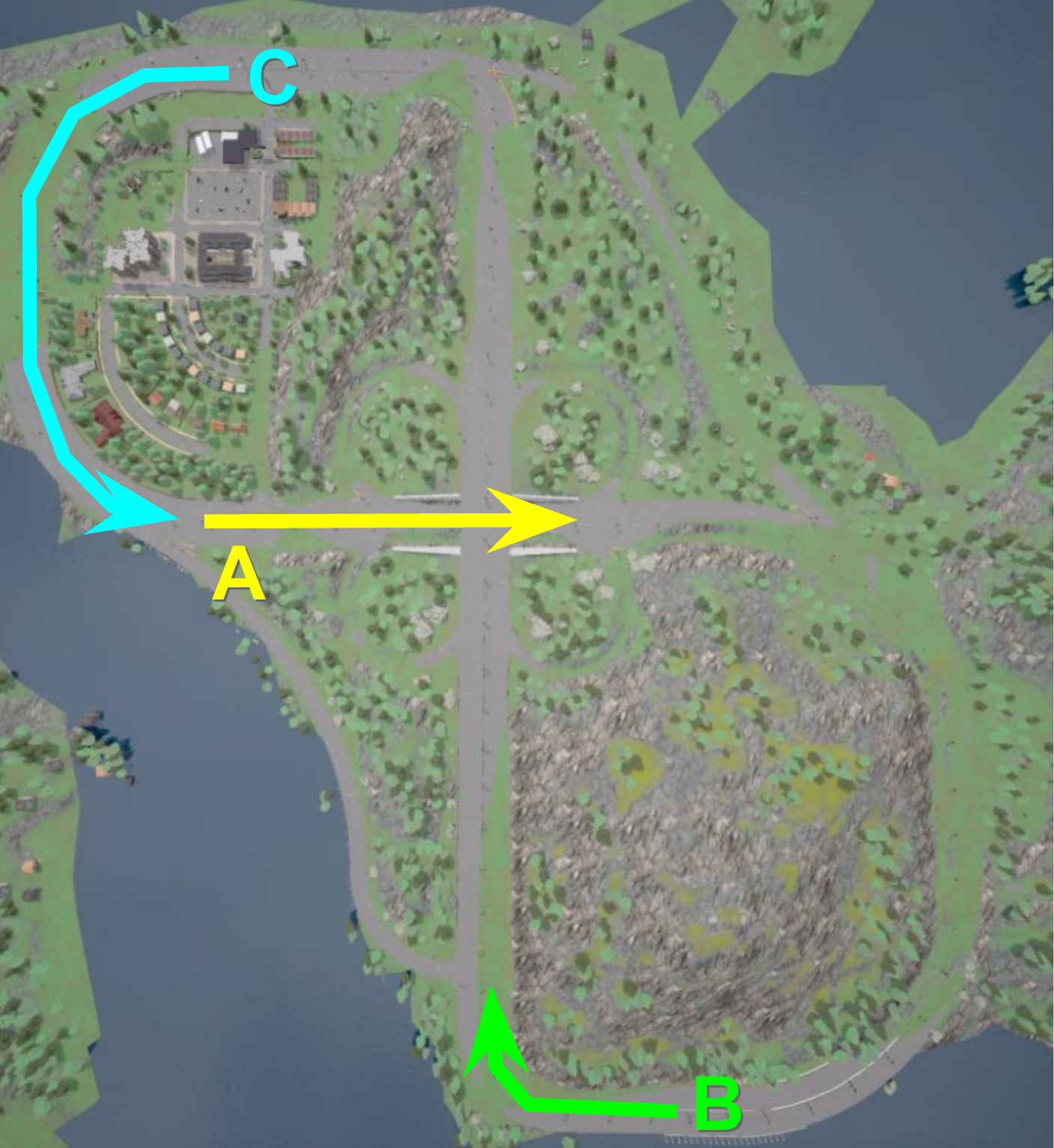}
   \includegraphics[width=.43\textwidth]{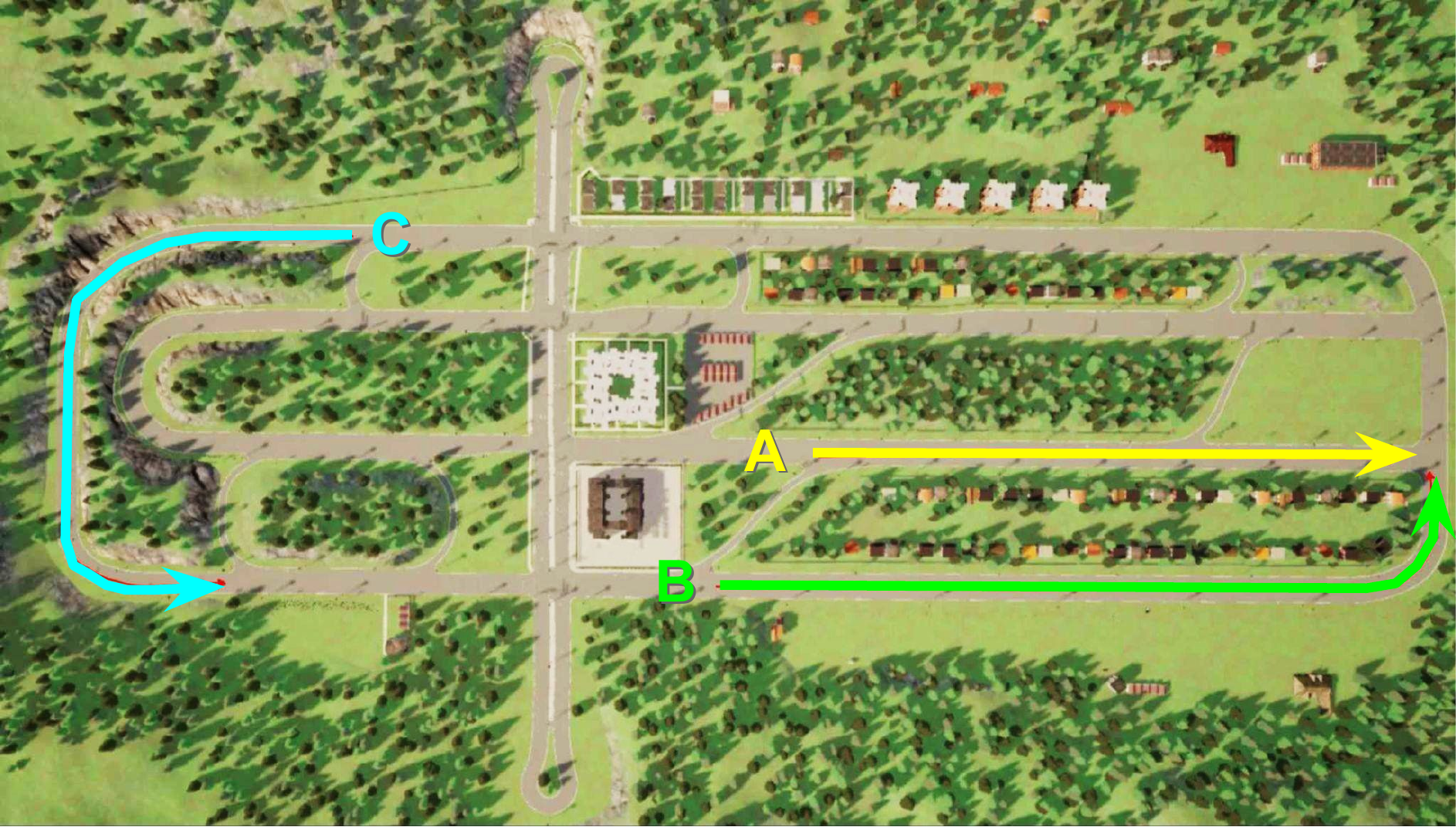}
\caption{Tracks used for testing in Town01, Town04, and Town06 from left to right. [A] shows the straight, [B] shows the one-turn, and [C] shows the two-turn track.}
\label{fig:tracks}
\end{figure*}

We trained our PML agent in Town06, which has long, many-lane highways with many highway entrances and exits. To conduct fair comparisons with a BC method, the BC agent was also trained in the same town. Then, these two agents were tested in Town01 (a small town with a river and several bridges) and Town04 (a small town embedded in the mountains with a loop highway). 
To assess their robustness, we conducted tests across four different lanes within each town. 
The test tracks used in these towns were selected randomly and are illustrated in Fig.~\ref{fig:tracks}. We selected three different driving scenarios in each town: straight, one-turn, and two-turn. They are labeled as [A], [B], and [C], respectively.

\subsubsection{PML}
For a task-agnostic exploration of the environment, 36,000 images in Town06, which enables rigorous testing across various lane configurations. To expand the dataset, images were augmented by flipping them along with their corresponding steering angles, resulting in a total of 72,000 image-steering pairs. A deliberate zigzag driving pattern was used during data collection to ensure a diverse and evenly distributed dataset covering the full range of steering angles from -1 (full left) to 1 (full right). This approach ensures the model learns the causal relationship between steering inputs and visual observations, improving its ability to generalize across different driving conditions. 

\begin{figure}
    \centering
    \includegraphics[width = 0.9\linewidth]{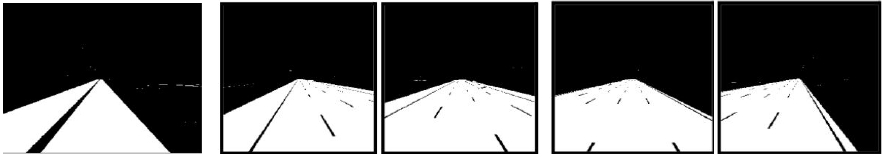}
    \caption{The examples of the \textit{preferences}. The first is for Town01 and Town02. The center two are for lanes 1 and 2 in Town04. The right two are lanes 3 and 4 in Town06.}
    \label{fig:preferences}
\end{figure}

Images used as \textit{preferences} for the PML agent are presented in Fig.~\ref{fig:preferences}. Since the road structure varies between towns, different \textit{preferences} were used to adapt the model accordingly. For benchmark experiments in Town01 and Town02, a single \textit{preference} image (the first image in Fig.~ \ref{fig:preferences}) was used for both towns, as their road structures were nearly identical.




To evaluate the effectiveness of our approach, we conducted two sets of experiments. First, we compared our PML-based method against a BC agent across different tracks in Town01, Town04, and Town06. Second, we tested our method using the CoRL2017 benchmark, which is widely used for evaluating autonomous driving policies. The standard CoRL2017 evaluation considers only RGB images, but since our method relies on \textit{road-only segmented images}, we added a semantic segmentation camera instead of RGB to extract relevant information. 

\subsubsection{Imitation Learning}
The BC agent was trained on a dataset collected from Town06, which is the same environment where the PML agent was trained and tested across all towns. Training BC agents in Town06 was particularly challenging due to the presence of multiple lanes, requiring a diverse dataset covering different lane positions. Additionally, the vehicle often veered too close to the curb when navigating turns. To ensure a fair comparison, the BC agent was trained exclusively for lateral control.
To capture high-quality driving behavior, we collected 61,000 samples, ensuring an approximately even distribution across different lanes in the simulator. Despite applying data augmentation techniques such as brightness adjustments and image flipping, the results remained unsatisfactory. To improve performance, we explicitly included flipped versions of the images in the dataset and applied a normalization strategy.

The collected data was normalized to balance the dataset across different steering angles. Specifically, we ensured that each steering bin contained a comparable number of images, preventing the model from being biased toward more frequently occurring steering values. Although this approach led to better driving behavior, the agent still exhibited difficulties in turns and struggled to recover when veering off the track. To address this, we collected an additional 32,000 samples, specifically demonstrating recovery maneuvers from different lanes.
After applying the normalization process, the final dataset consisted of 47,313 images paired with steering angles, ensuring a well-distributed representation of different driving scenarios. 
 All models are implemented on the TensorFlow framework. The lambda Workstation having two NVIDIA RTX 6000 was used for training the images of semantic segmentation and RGB images, and NVIDIA GeForce RTX 4060 was used for training grayscale images and doing the tests in the CARLA simulator.
 
\section{Results}

\subsection{Performance Comparison}

PML proved to be a robust and adaptable agent for autonomous driving, as evidenced by the comprehensive analysis presented in Table \ref{table:performance-comparison}. In the table, the deviation is the average value of the Euclidean distance between the vehicle's positions and the corresponding waypoints, and the success rate shows, across different tracks, how many tracks were successfully completed. For each track, four different experiments were conducted, and the success rate was computed based on the success of these experiments. The deviations should be considered bearing in mind that the whole lane in Town01 has a width of 4 meters, and Town04 and Town06 have a width of 3.5 meters.
Table \ref{table:performance-comparison} presents the performance metrics of PML and BC agents across various tasks in different urban environments. The performance is evaluated using two key metrics: average deviation and success rate.
The results indicate that the PML agent consistently outperforms the BC agent across various tasks and environments, achieving higher success rates and lower deviations in most cases. In Town01, PML demonstrated perfect performance with a 100\% success rate in all tasks, whereas BC struggled. In Town04, PML maintained strong performance across tasks, with BC showing comparable results only in the Straight and Two Turns task. In Town06, both agents performed similarly in terms of success rate, but PML exhibited lower deviation, indicating more precise trajectory tracking. Overall, PML proved to be a more robust and adaptive approach, particularly in challenging scenarios.

\begin{table}
\centering
\caption{Performance comparison}
\label{table:performance-comparison}
\begin{tabular}{lllcc}
\hline
Town   & Agent   & Task       & \multicolumn{1}{c}{Avg. Dev. (↓)} & \multicolumn{1}{c}{Success (\% ↑)} \\
\hline
\multirow{8}{*}{Town01} & \multirow{4}{*}{PML} & Straight   & 0.4080  & 100.00  \\
       &        & One-Turn   & 0.5400  & 100.00 \\
       &        & Two Turns  & 0.4860  & 100.00 \\
       \cmidrule(lr){3-5}
       &        & Overall    & \textbf{0.4780}  & \textbf{100.00} \\
       \cmidrule(lr){2-5}
       & \multirow{4}{*}{BC} & Straight   & 0.9780  & 0.00 \\
       &        & One-Turn   & 1.2170  & 0.00 \\
       &        & Two Turns  & 3.4180  & 25.00 \\
       \cmidrule(lr){3-5}
       &        & Overall    & 1.8710  & 8.33 \\
\hline
\multirow{8}{*}{Town04} & \multirow{4}{*}{PML} & Straight   & 0.8120  & 100.00 \\
       &        & One-Turn   & 0.9440  & 100.00 \\
       &        & Two Turns  & 0.8230  & 100.00 \\
       \cmidrule(lr){3-5}
       &        & Overall    & \textbf{0.8597}  & \textbf{100.00} \\
       \cmidrule(lr){2-5}
       & \multirow{4}{*}{BC} & Straight   & 0.6880  & 100.00 \\
       &        & One-Turn   & 1.7140  & 0.00 \\
       &        & Two Turns  & 0.8750  & 100.00 \\
       \cmidrule(lr){3-5}
       &        & Overall    & 1.0923  & 66.67 \\
\hline
\multirow{8}{*}{Town06} & \multirow{4}{*}{PML} & Straight   & 0.7010  & 75.00 \\
       &        & One-Turn   & 0.5730  & 100.00 \\
       &        & Two Turns  & 0.6460  & 100.00 \\
       \cmidrule(lr){3-5}
       &        & Overall    & \textbf{0.6400}  & \textbf{91.67} \\
       \cmidrule(lr){2-5}
       & \multirow{4}{*}{BC} & Straight   & 1.1630  & 75.00 \\
       &        & One-Turn   & 1.7840  & 100.00 \\
       &        & Two Turns  & 1.4790  & 100.00 \\
       \cmidrule(lr){3-5}
       &        & Overall    & 1.4753  & \textbf{91.67} \\
\hline
\end{tabular}
\end{table}

\subsection{CoRL2017 Benchmark} 
We also compared our results with various methods, including Modular Pipeline (MP), Imitation Learning (IL), and Reinforcement Learning (RL), as implemented in the CARLA simulator \cite{CARLA}. Additionally, we included benchmark results from CIRL \cite{Cirl}, CAL \cite{Conditional-affordance-learning-for-driving-in-urban-environments}, and LBC \cite{Learning-by-cheating}, as shown in Table \ref{table:corl2017-benchmark}. For these studies, Town01 was used for training, while Town02 served as the testing environment. Note that both towns were not seen to our PML agent in the training stage. 

\begin{table}
\centering
\setlength{\tabcolsep}{0.65em} 
\caption{Success rate result in CoRL2017 benchmark.}
\label{table:corl2017-benchmark}
\begin{tabular}{llllllll} 
\toprule
              & MP& IL& RL& CAL& CIRL& LBC& PML\\ 
               & \cite{CARLA}& \cite{CARLA}  & \cite{CARLA} & \cite{Conditional-affordance-learning-for-driving-in-urban-environments} & \cite{Cirl} & \cite{Learning-by-cheating}& (Ours)\\
\midrule
Town01-Train & 98 & 95 & 89 & 100 & 98   & 100 & 96 (Test)\\
Town02-Test  & 92 & 97 & 74 & 93  & 100  & 100 & 92 (Test)\\
\bottomrule
\end{tabular}
\end{table}

Our PML with AIF method outperforms IL and RL in the straight driving task in Town01, though it is slightly outperformed by the other approaches (See Table \ref{table:corl2017-benchmark}). However, it is important to note that, unlike other methods, we did not use any training data from Town01, which gives our approach a strong generalization advantage. In Town02, our agent demonstrates driving performance comparable to CAL and MP and surpasses RL in the straight task.

Although CIRL and LBC achieve better results, our method offers a significantly simpler training pipeline. LBC employs a two-stage training process, making it more computationally demanding and resource-intensive. Similarly, CIRL integrates a dual-stage learning process that adds complexity in both implementation and hyperparameter tuning. The need to balance IL and RL in CIRL introduces additional challenges, such as overfitting to demonstration data or inadequate generalization from RL. Overall, our PML with AIF-based approach delivers competitive results, highlighting its potential as an alternative to more complex learning frameworks.

\section{Conclusion}
In this paper, we introduced a novel PML-based method for controlling an autonomous vehicle, leveraging the FEP to dynamically adapt to different road types, including two-lane and multi-lane environments. Our method significantly enhances the adaptability and performance of HAVs, achieving comparable results to more complex baseline methods in the CARLA benchmark. Our results demonstrate that the PML agent, despite its simplicity, performs robustly across various urban environments and driving tasks. It consistently outperforms BC in terms of lower average deviations and higher success rates in simulated environments. 

Future work will aim to enhance the capabilities of our PML-based agent by incorporating longitudinal control (throttle and brake) and addressing more complex tasks such as lane changing, handling dynamic objects in the driving scene, and adhering to traffic rules. Additionally, we aim to validate our approach using real-world data and explore its application to actual autonomous driving scenarios. 
While this study focused on the lane-keeping task, our approach can be extended to more complex maneuvers, such as lane-changing. 

\bibliography{main}

\end{document}